\documentclass{article}

\usepackage{lipsum}

\usepackage[final, nonatbib]{neurips_2024}

\usepackage[utf8]{inputenc} 
\usepackage[T1]{fontenc}    
\usepackage{hyperref}       
\usepackage{url}            
\usepackage{booktabs}       
\usepackage{amsfonts}       
\usepackage{nicefrac}       
\usepackage{microtype}      
\usepackage{xcolor}         
\usepackage{amsmath} 
\usepackage{empheq}
\usepackage{bbm}
\usepackage{breqn}
\usepackage{amsmath,amsfonts,amssymb,amsthm}
\usepackage{newpxtext} 
\usepackage[framemethod=tikz]{mdframed}

\usepackage[
backend=biber,
style=numeric,
]{biblatex}
\title{\
Leveraging Intermediate Neural Collapse with Simplex ETFs for Efficient Deep Neural Networks
}

\author{%
  Emily Liu \\
  Department of Electrical Engineering and Computer Science\\
  Massachusetts Institute of Technology\\
  Cambridge, MA 02139 \\
  \texttt{emizfliu@mit.edu} \\
}

\begin{document}

\maketitle

\begin{abstract}
  Neural collapse is a phenomenon observed during the terminal phase of neural network training, characterized by the convergence of network activations, class means, and linear classifier weights to a simplex equiangular tight frame (ETF), a configuration of vectors that maximizes mutual distance within a subspace. This phenomenon has been linked to improved interpretability, robustness, and generalization in neural networks. However, its potential to guide neural network training and regularization remains underexplored. Previous research has demonstrated that constraining the final layer of a neural network to a simplex ETF can reduce the number of trainable parameters without sacrificing model accuracy. Furthermore, deep fully connected networks exhibit neural collapse not only in the final layer but across all layers beyond a specific effective depth. Using these insights, we propose two novel training approaches: \emph{Adaptive-ETF}, a generalized framework that enforces simplex ETF constraints on all layers beyond the effective depth, and \emph{ETF-Transformer}, which applies simplex ETF constraints to the feedforward layers within transformer blocks. We show that these approaches achieve training and testing performance comparable to those of their baseline counterparts while significantly reducing the number of learnable parameters.
\end{abstract}

\section{Introduction}
In the past decade, steady advances in machine learning have propelled deep neural networks to achieve superhuman performance on a wide range of challenging tasks, including image classification, speech recognition, and natural language processing. Despite these successes, the intricate structure of deep neural networks often renders them opaque and difficult to interpret, leading to their characterization as ``black boxes." However, recent research has revealed that neural networks can exhibit mathematically simple and elegant structures during training. One such phenomenon, known as neural collapse, has garnered significant attention for its implications in understanding and optimizing neural network behavior.

Neural collapse (NC) describes the convergence of learned features within a neural network to highly structured geometric patterns during the training process. Specifically, NC is characterized by alignment of features with their corresponding class means, convergence of class means to linear classifier weights, and the arrangement of these means and weights as a simplex equiangular tight frame (ETF), a configuration that maximizes pairwise distances within a subspace. This phenomenon, first documented by Papyan et al. \cite{nc}, occurs during the terminal phase of training (TPT), a stage where the network achieves zero training error, and the loss function asymptotically approaches zero. Understanding the neural collapse phenomenon holds promise for designing neural networks with improved interpretability, robustness, and generalization capabilities.

\paragraph{Contribution}

Previous work has largely focused on exploring neural collapse from a theoretical lens, focusing on its characterization and providing mathematical justifications for its emergence. More recently, tscope of neural collapse has been expanded to include intermediate outputs within neural networks, revealing connections to generalization bounds \cite{ed}.
Despite these advances, the practical implications of using simplex ETF during training, particularly in relation to neural collapse, remain largely unexplored.
In this work, we address this gap by demonstrating the viability of employing intermediate-layer simplex ETFs to reduce memory usage during training. Furthermore, we extend this approach to transformer architectures, showing that while neural collapse is not directly observed in the fully connected layers of transformer blocks, replacing these layers with simplex ETFs leads to faster training with negligible impact on prediction accuracy. These contributions underscore the potential of integrating neural collapse principles into practical training frameworks for both fully connected and transformer-based architectures.



\section{Background: Neural Collapse}
The task is a multiclass classification problem. We consider a distribution $P$ over samples $(x, y) \in \mathcal{X} \times \mathcal{Y}_C$ where $\mathcal{X} \subset \mathbb{R}^d$ and $\mathcal{Y}_C = [C]$, and a deep neural network, which consists of a feature learner (all previous layers) $h(x) \in \mathbb{R}^k$ followed by a linear classifier of weights $W$ and bias $b$; we can therefore consider the entire network itself as $g(x) = W h(x) + b$. The regularized empirical objective function over i.i.d. observed dataset $S$ using cross entropy loss $\ell$ is expressed as
\begin{align*}
\mathcal{L}_S^\lambda (g_W) := \frac{1}{|S|} \sum_{(x_i, y_i) \in S} \ell (g_W(x_i), y_i) + \lambda \|W\|_2^2.
\end{align*}

There are four interconnected neural collapse conditions, stated below.\cite{nc}

\subsection{(NC1) Variability collapse} The within-class variance of the features $h$ go to zero, which means that all features collapse to class means.
\begin{align*}
\Sigma_W = \mathbb{E}_{i, c}[(h_{i, c} - \mu_c)(h_{i, c} - \mu_c)^\intercal] \rightarrow 0
\end{align*}

\subsection{(NC2) Convergence to Simplex Equiangular Tight Frame (ETF)} Over training, the vectors of the class means (in matrix $\dot{M}$) converge to having equal length, having the same angle between any pair of class means, and being maximally pairwise distant from each other.
\begin{align*}
    |\|\mu_c - \mu_G\|_2 - \|\mu_{c'} - \mu_G\|_2| &\rightarrow 0,\\
    \biggl<\frac{\mu_c - \mu_G}{\|\mu_c - \mu_G\|_2}, \frac{\mu_{c'} - \mu_G}{\|\mu_{c'} - \mu_G\|_2} \biggr> &\rightarrow \frac{C}{C-1} \delta_{c, c'} - \frac{1}{C-1} \quad\quad \forall c, c' 
\end{align*}
This is equivalent to the simplex equiangular tight frame (ETF) a well-studied structure in mathematics. The simplex ETF is described by the columns of the matrix
\begin{align*}
\dot{M} = \sqrt{\frac{C}{C-1}} P \left(I_C - \frac{1}{C} \mathbb{1}_C \mathbb{1}_C^\intercal\right)
\end{align*}
where $P$ forms the first $C$ columns of an identity matrix of size $k$. \cite{etf}
\subsection{(NC3) Convergence to self-duality} The class means and weights occupy dual vector spaces. In neural collapse, they converge to each other up to a rescaling.
\begin{align*}
    \biggl\|\frac{W^\intercal}{\|W\|_F} - \frac{\dot{M}^\intercal}{\dot{M}\|_F}\biggr\| \rightarrow 0
\end{align*}
\subsection{(NC4) Simplfication to nearest class center (NCC)} For any data point, the classifier converges to choosing the class has the nearest training class mean in Euclidean distance.
\begin{align*}
    \arg\max_{c'} <w_{c'}, h> + b_{c'} \rightarrow \arg\min_{c'} \|h - \mu_{c'} \|_2
\end{align*}
Because the NC4 metric is most relevant to neural network intermediates, and because the equivalence of neural collapse conditions has been shown in \cite{nc}, we use NCC accuracy to measure the degree of neural collapse for our experiments.

\section{Related work: Simplex ETF and Effective Depth}
\paragraph{Effective Depth}
Galanti et. al. \cite{ed} defines the notion of \textit{effective depth} $L_0$ of a deep neural network as the minimum depth at which a deep neural network achieves NCC separability (NCC error $\leq \epsilon$). If the network does not achieve NCC separability, then $L_0 = L$. The minimal-depth hypothesis \cite{ed} states that there must exist some depth $L_0 \geq 1$ such that any network with depth greater than $L_0$ also achieves NCC separability. It follows that the NCC accuracy of layer $L$ must be greater than or equal to the NCC accuracy of layer $L-1$ for all layers $L \geq L_0$.
Less strictly, $\epsilon$-effective depth denotes the minimal depth of a network at which the NCC error is less than or equal to $\epsilon$. $\epsilon$-effective depth is the metric we use when evaluating effective depths of the neural networks in Section \ref{Experiments}.
\paragraph{Simplex ETF for neural network training}
By NC2 and NC3, class means and linear classifiers both converge to a simplex ETF. Recent work by Zhu et. al. \cite{etf} has shown that if a network has exhibited neural collapse, then setting the final layer to be a simplex ETF allows for reduced parameters in the network while maintaining the same degree of accuracy as an otherwise equivalent model with a fully trainable final layer. This enables significantly reduced training-time memory usage.

\paragraph{Combining Effective Depth and Simplex ETF}
In this paper, we propose a training scheme that utilizes the simplex ETF framework for all layers past the $\epsilon$-minimal depth of a given neural network. It has been previously demonstrated that a network where $L_0 \geq L$ with the final layer fixed to a simplex ETF achieves the same accuracy as an unconstrained network with the same architecture, using far less memory. Results from Galanti et. al. \cite{ed} show that overparameterized neural networks achieve NCC separability in not only the last layer, but in all layers between $L_0$ and the last layer. It follows that we can further extend the simplex ETF constraint to intermediate layers beyond the effective depth with no detriment to training. This demonstrates not only a method of training that saves on memory usage and number of learnable parameters, but also a link between NC2, NC3, and NC4 conditions that extends beyond the final layer of the neural network.

\section{Experiments}\label{Experiments}
\paragraph{Training}
We use the standardized Fashion-MNIST image dataset \cite{2017arXiv170807747X}. We use a cross-entropy loss function and train for 300 epochs with stochastic gradient decent, using weight decay 5e-4, momentum 0.9, and a learning rate schedule with initial learning rate 0.01, with decay factor 0.1 at epochs 60, 120, and 160. We use batch size 128 for all models.

\subsection{Intermediate Simplex ETF in Multilayered Perceptrons}
\begin{figure}[h]
    \centering
    \includegraphics[scale=0.25]{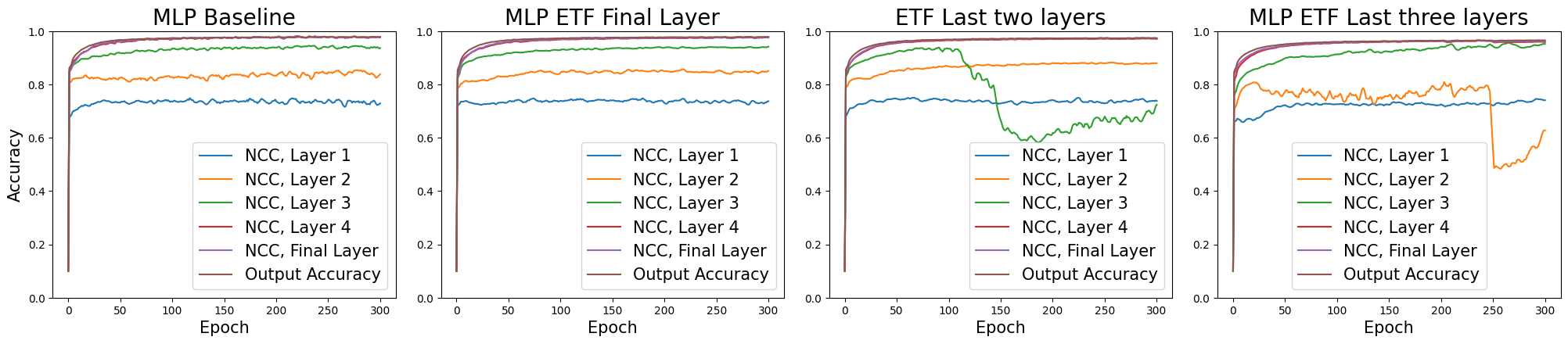}
    \caption{Training progress on baseline MLP compared to replacing the final layer, final two layers, and final three layers of the neural network with simplex ETFs.}
    \label{fig:etfs}
\end{figure}

We train and evaluate a 5-layer feedforward multi-layered perceptron with intermediate layers of width 128, ReLU activation, and batch normalization on intermediate layers.
Figure \ref{fig:etfs} shows the compares of training the neural network without ETF constraints (baseline) to training the neural network while replacing the last 1, 2, and 3 layers with Simplex ETFs. The baseline results indicate that the network has an effective depth $L_0$ of 4. We observe that setting all layers past $L_0$ to simplex ETFs does not impact the train or test accuracy of the network.

Indeed, as can be seen in Figure \ref{fig:etfs}, setting final layers to ETFs does not affect the final accuracy of the neural network, and the modified neural networks still have effective depths of 4. However, the behavior of the intermediate layers under more constrained conditions (last two or three layers ETF) differs from baseline. In addition to the final training and validation accuracy being lowered (Table \ref{tab:mlp_results}), the last fully-connected layer before ETF layers exhibits lower NCC separability than other layers. 

In both instances of over-constrained neural networks, the decrease in NCC separability is marked by a noticeable ``phase change"— the NCC accuracy starts off high but decreases dramatically after the network accuracy improves past a certain point in training. This change does not affect the NCC separability of earlier layers, nor does it significantly affect train or test accuracy. Although this form of mode collapse is only observed in a heavily constrained neural network, it may hint towards underlying dynamics in neural network training that can affect feature representation and separability even in less constrained models. This suggests that similar mechanisms of mode collapse might occur subtly in standard networks, potentially impacting generalization and the interpretability of learned representations. Understanding this phenomenon could lead to better network designs and training strategies that preserve feature separability throughout the network layers.

\begin{table}[]
    \centering
    \begin{tabular}{cccc}
    \hline
        Fashion-MNIST  & \# parameters & Train Accuracy & Test Accuracy \\
        \hline
        Baseline MLP & 152586 & 98.02\% & 89.58\% \\
        Last layer ETF & 151306 & 98.13\% & 89.44\% \\
        Last two layers ETF & 134922 & 97.59\% & 89.20\% \\
        Last three layers ETF & 118538 & 96.87\% & 88.50\%\\
        Adaptive ETF & - & 98.09\% & 89.38\% \\
    \hline
        \\
    \end{tabular}
    \caption{Best training and validation accuracy attained from MLP models.}
    \label{tab:mlp_results}
\end{table}
\begin{table}[]
    \centering
    \begin{tabular}{cccc}
        \hline
        Fashion-MNIST & \# Parameters & Train Accuracy & Test Accuracy \\
        \hline
        Baseline ViT & 9491978 & 93.91\% & 89.80\%\\
        Last layer ETF & 9486858 & 93.89\% & 90.03\% \\
        Last layer + Last Block FC ETF & 7389706 & 98.89\% & 90.06\%  \\
        Last layer + Last 2 Blocks FC ETF & 5292554 & 93.51\% & 89.97\% \\
        ETF-Transformer & 3195402 & 93.23 \% & 89.78\% \\
        \hline
        \\
    \end{tabular}
    \caption{Best training and validation accuracy attained from Vi-T model.}
    \label{tab:vit_results}
\end{table}

\begin{figure}[h]
    \centering
    \includegraphics[scale=0.3]{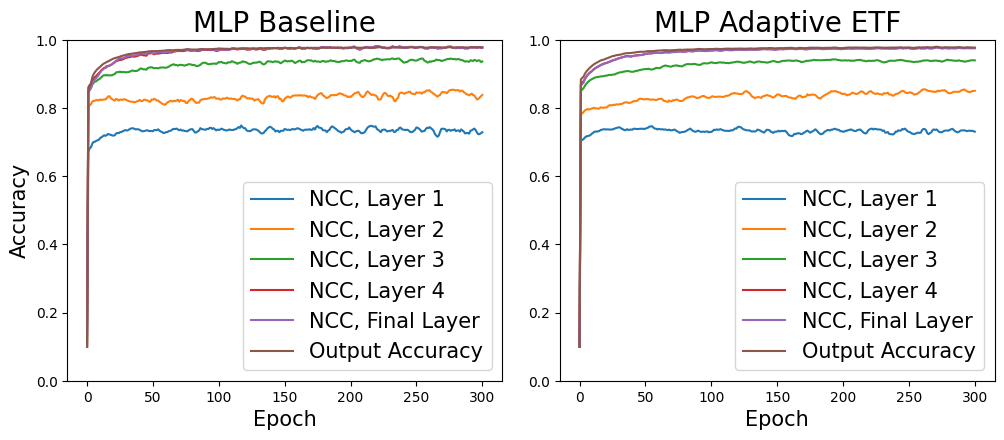}
    \caption{Training progress on baseline network compared to replacing feedforward layers with simplex ETFs beyond the effective depth.}
    \label{fig:adaptive}
\end{figure}

\paragraph{New training scheme: Adaptive ETF}
Using results from training ETF-constrained MLPs, we observe that replaying layers beyond a fully-connected network's effective depth with simplex ETFs does not incur a penalty to train or test accuracy. This suggests a new training strategy, \emph{Adaptive ETF}, that can be employed to greatly reduce memory usage in overparameterized neural networks. Given parameters $\epsilon$, we fix a given layer of the neural network to a simplex ETF if its NCC error falls under $\epsilon$. Figure \ref{fig:adaptive} demonstrates this technique in use with $\epsilon = 0.1$. We clearly see that the model achieves the same train accuracy, test accuracy, and effective depth as a baseline network, demonstrating Adaptive ETF as a feasible training technique.

\subsection{Extension to Transformer Architectures: Introducing ETF-Transformer}

\begin{figure}[h]
    \centering
    \includegraphics[scale=0.27]{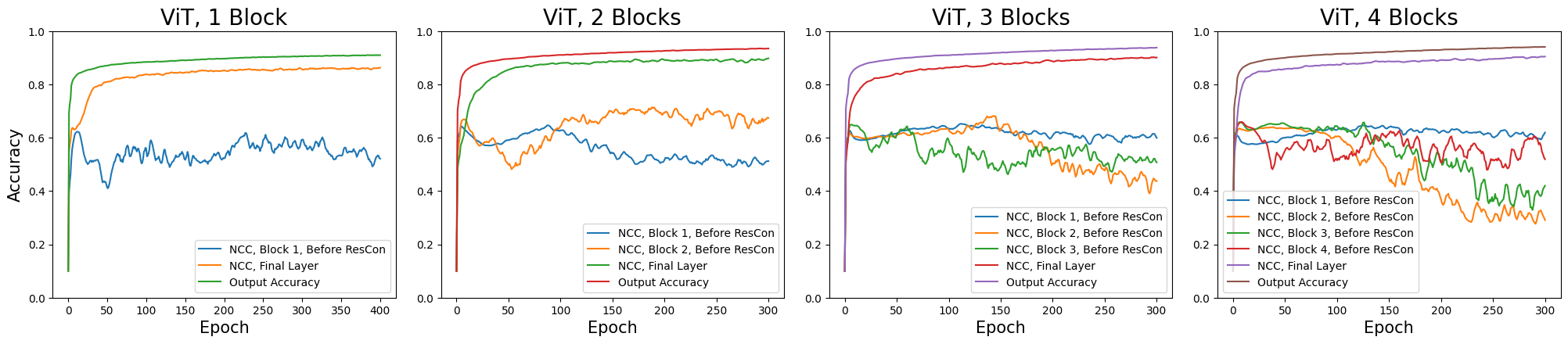}
    \includegraphics[scale=0.27]{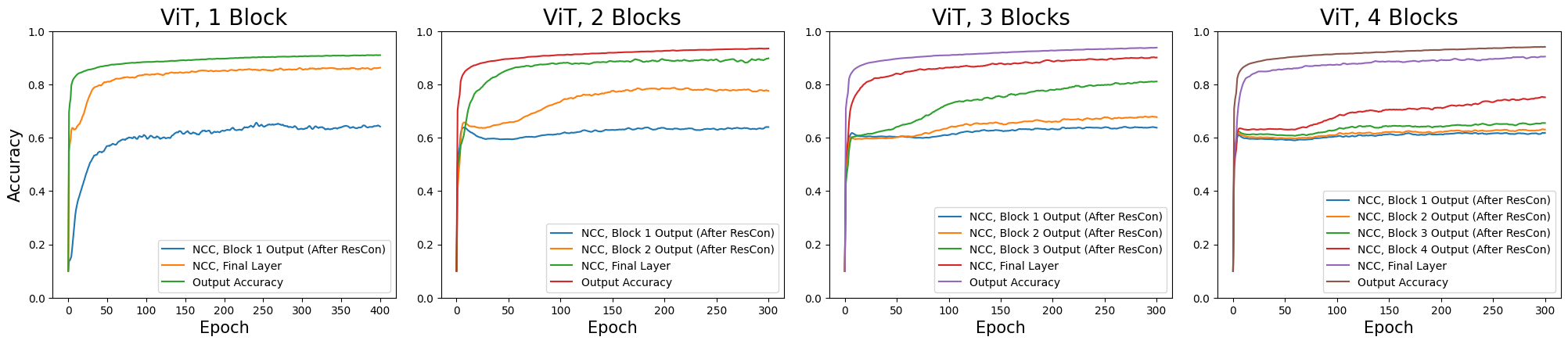}
    \caption{Comparison of training progress on vision transformers with 1 block, 2 blocks, 3 blocks, and 4 blocks. }
    \label{fig:tformerlayers}
\end{figure}
In the previous sections, we have demonstrated the efficacy of using simplex ETFs to replace weights exhibiting neural collapse with fully-connected networks. However, many real-world tasks make use of more complicated models, such as transformers. We seek to translate as many of our insights as possible to these complicated models, even if the neural collapse phenomenon has not been as well-documented in these contexts. 

To this end, we repeat our experiments on a visual transformer (ViT) model \cite{dosovitskiy2021an}. Each transformer block has 8 heads, an attention dimension of 64, and 2048 hidden units in the fully connected layer. We will investigate the effect of constraining the fully-connected layers in both the output and the transformer blocks to simplex ETFs.

\begin{figure}[h]
    \centering
    \includegraphics[scale=0.28]{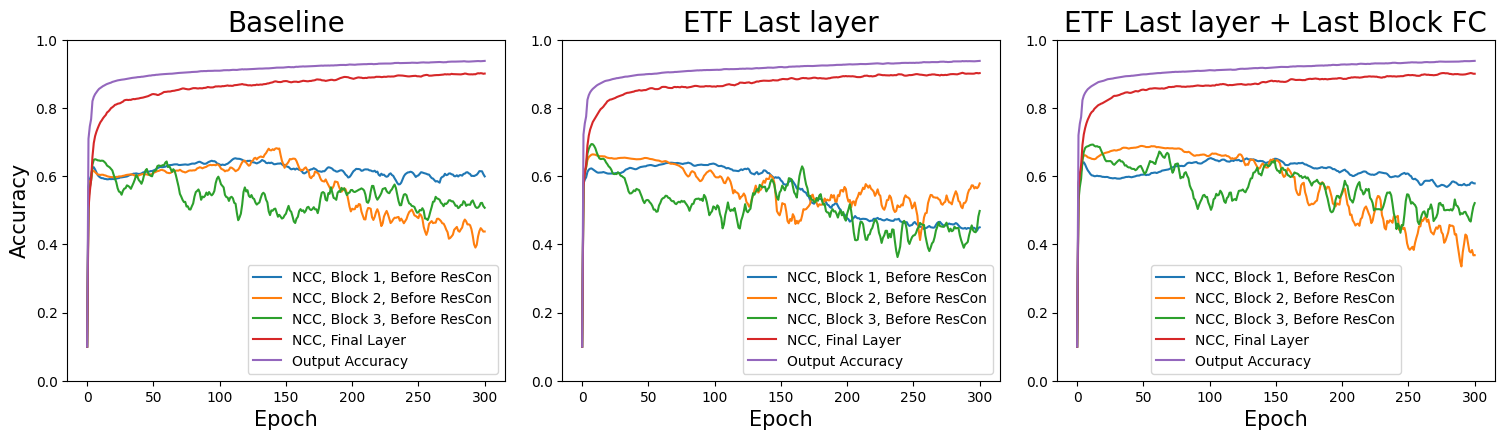}
    \includegraphics[scale=0.28]{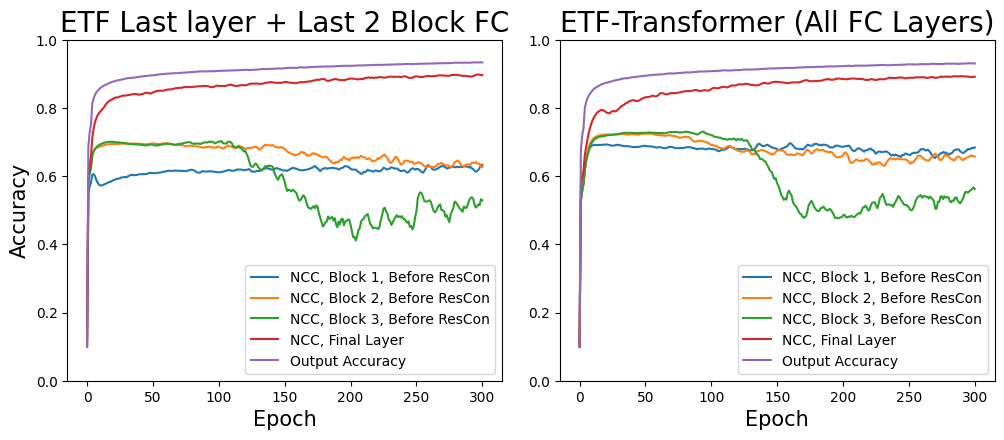}
    \caption{NCC for FC layers in transformer ETF architectures.}
    \label{fig:tformeretfs}
\end{figure}

As seen in Figure \ref{fig:tformerlayers}, the effective depth of the ViT model is always equal to one more than the number of blocks. That is, the notion of effective depth in MLPs does not carry over cleanly to transformers. NCC accuracy for intermediates after the residual connection is greater than for intermediates before the residual connection. This is consistent with the idea that NCC separability is achieved as a result of optimization rather than specifics of the neural network. However, it is important to note that the direct output of the FC layers (before residual connection) in the transformer architecture likely plays a similar role to the last non-ETF layer in the MLP network in Figure \ref{fig:etfs}.

At the same time, experimental results show that constraining fully-connected layers to simplex ETFs does not significantly impact performance: Table \ref{tab:vit_results} and Figure \ref{fig:tformeretfs} show the accuracy and intermediate NCC results obtained on training a ViT model with 3 transformer blocks, constraining first the final layer, then the final layer plus the transformer FC layers in the last block, then the final layer plus the transformer FC layers in the last two blocks, and finally setting all FC layers outside of multi-head self attention to ETFs. Because we are able to set ETFs on all transformer blocks, we generalize this last training scheme as the \emph{ETF-Transformer} approach. We note that due to the expressivity of the non-MLP components of the transformer model, the introduction of ETFs into ViT only slightly decrease the training accuracy and do not affect generalization.

\section{Conclusion and Future Work}
In this paper, we demonstrate the feasibility of using equiangular tight frames in training neural networks. Experiments show that when later layers $(L \geq L_0)$ are fixed to ETF, the network trains to the same accuracy as an unmodified network. This suggests a novel training scheme, Adaptive ETF, in which layers of the neural network are set to simplex ETFs once they achieve NCC separability. We demonstrate that Adaptive ETF achieves similar performance to an unmodified neural network on the Fashion-MNIST dataset.

We also demonstrate the strengths and limitations of carrying over the notion of simplex ETFs to transformers, using the ViT model. Even though neural collapse conditions are not as prevalent as in MLPs, we observe that setting feedforward layers to simplex ETFs in transformer blocks still results in a model that achieves similar accuracy to baseline, 

This investigation opens up several new directions of research. First, the Adaptive-ETF technique and ETF-Transformer architecture can be applied to larger classification datasets, as well as datasets in different modalities such as natural language. Additionally, our experiments demonstrate that NC2 and NC3 can also be applied to intermediate layers, meaning that it is worthwile to revisit NCC-based generalization bounds \cite{ed} from the perspective of NC2 or NC3. Finally, our work yielded unexpected results when fixing early layers $(L < L_0)$ to simplex ETFs: the training accuracy decreased but the test accuracy stayed the same. This suggests an interesting avenue of exploration for using simplex ETFs in neural network regularization.

\nocite{10.5555/3295222.3295349}
\nocite{alain2017understanding-key}
\printbibliography


\end{document}